\documentclass[twocolumn,english]{IEEEtran}
\usepackage[T1]{fontenc}
\usepackage{float}
\usepackage{amsmath}
\usepackage{amsthm}
\usepackage{amssymb}
\usepackage{graphicx}
\PassOptionsToPackage{normalem}{ulem}
\usepackage{ulem}

\makeatletter

\floatstyle{ruled}
\newfloat{algorithm}{tbp}{loa}
\providecommand{\algorithmname}{Algorithm}
\floatname{algorithm}{\protect\algorithmname}

\theoremstyle{plain}
\newtheorem{thm}{\protect\theoremname}
\theoremstyle{definition}
\newtheorem{defn}[thm]{\protect\definitionname}
\theoremstyle{definition}
\newtheorem{example}[thm]{\protect\examplename}

\usepackage{balance}

\IEEEoverridecommandlockouts

\newcommand{\bmat}{\left[\begin{array}}
\newcommand{\emat}{\end{array}\right]}

\usepackage{dsfont}

\usepackage{algorithm,algpseudocode}

\makeatother

\usepackage{babel}
\providecommand{\definitionname}{Definition}
\providecommand{\examplename}{Example}
\providecommand{\theoremname}{Theorem}

\begin{document}

\title{Atypicality for Heart Rate Variability Using a Pattern-Tree Weighting
Method}

\author{Elyas Sabeti, \emph{Member, IEEE} and Anders H{\o}st-Madsen, \emph{Fellow,
IEEE}}

\maketitle
\global\long\def\cov{\mathrm{cov}}

\begin{abstract}
Heart rate variability (HRV) is a vital measure of the autonomic nervous
system functionality and a key indicator of cardiovascular condition.
This paper proposes a novel method, called pattern tree which is an
extension of Willem's context tree to real-valued data, to investigate
HRV via an atypicality framework. In a previous paper atypicality
was developed as method for mining and discovery in \textquotedblleft Big
Data,\textquotedblright{} which requires a universal approach. Using
the proposed pattern tree as a universal source coder in this framework
led to discovery of arrhythmias and unknown patterns in HRV Holter
Monitoring. 
\end{abstract}

\vspace{-0.2in}

\section{Introduction}

Information theory is generally a theory of typicality. For example,
compressing data using the Asymptotic Equipartion Property (AEP) can
be done by throwing away all sequences that are not typical. Our perspective
in this paper and prior work \cite{Host13ITW,Host15ISITl,HostSabetiWalton16IEEEtrans,MadsenSabeti2016BigData,Sabeti2015GLobSIT,Sabeti2016ExpoFam,Sabeti2016Image,SabetiHost17IEEEtrans,Sabeti2017ISITpredictive}
is that the value of data lies not in these typical sequence, but
in the atypical sequences. Take art: the truly valuable paintings
are those that are rare and atypical. Take online collections of photos,
such as Flickr.com,  the photos that of interest are those that are
unique. They are atypical. Of course, as opposed to 'interestingness,'
an atypicality criterion will find photos that are both uniquely good
and uniquely bad, there is no value judgment. A similar example can
be investing: extraordinary gains can be obtained only by investing
in atypical stocks, yet that can also lead to total ruin. Atypicality
\cite{HostSabetiWalton16IEEEtrans} is defined by

\vspace{-0.1in}

\begin{defn}
\label{atypdef.thm}\emph{A sequence is atypical if it can be described
(coded) with fewer bits in itself rather than using the (optimum)
code for typical sequences}.
\end{defn}
\vspace{-0.1in}

In prior work, this framework has been used for data discovery \cite{Host13ITW,Host15ISITl,HostSabetiWalton16IEEEtrans,MadsenSabeti2016BigData,Sabeti2015GLobSIT,Sabeti2016ExpoFam,Sabeti2016Image,SabetiHost17IEEEtrans,Sabeti2017ISITpredictive}.
Our aim with atypicality theory is to find \emph{\uline{'unknown
unknowns}}' \cite{Rumsfeld}. To encode data in \emph{itself}, we
require a universal source coder. In our discrete case atypicality
papers we used Willems' Context-Tree Weighting (CTW) as an universal
source coder which requires data binarization.

With the aforementioned atypicality purpose, in this paper we are
extending CTW method to real-valued data by introducing the pattern-tree
weighting (PTW) method with finite maximum memory depth as binary-structured
tree in which data samples are partitioned by a binary \emph{pattern}
achieved by comparing consecutive samples and the probability estimation
in each node is done by predictive estimators such as predictive minimum
description length (MDL). 

The principal reason for introducing a new method such as PTW is its
strength in pattern mining of real-valued data which has enormous
applications in data science. For instance, the patterns in Heart
Rate Variability (HRV) are informative and symptomatic of heart diseases;
however their detailed recognition in Holter Monitoring is impractical
for cardiologists, due to huge amount of data at hand, i.e., ``Big
Data.'' We first discovered these HRV patterns in \cite{Host13ITW},
but because of the data binarization many patterns of real-valued
HRV were not captured. This issue is circumvented by PTW.\vspace{-0.2in}

\section{\label{sec:Predictive-MDL}Predictive MDL}

Consider the family of parametric models $\mathcal{P}=\left\{ P_{\boldsymbol{\theta}}\left(\mathbf{x}_{n}\right):\:\boldsymbol{\theta}\in\Theta\subset\mathcal{\mathbb{R}}^{k}\right\} $.
Rissanen \cite{Rissanen86} defined predictive MDL (we call it ordinary
predictive MDL) by\vspace{-0.1in}

\[
L(\mathbf{x}_{n})=-\sum_{i=1}^{n-1}\log P_{\boldsymbol{\hat{\theta}}(\mathbf{x}_{i})}(x_{i+1}|\mathbf{x}_{i})
\]
in which using the already observed data $\mathbf{x}_{i}$, the parameters
of the model $\boldsymbol{\hat{\theta}}(\mathbf{x}_{i})$ are estimated
and they're used to estimate the probability of the next sample $P_{\boldsymbol{\hat{\theta}}(\mathbf{x}_{i})}^{op}(x_{i+1}|\mathbf{x}_{i})$.
In our companion paper \cite{Sabeti2017ISITpredictive} we have shown
that the ordinary predictive MDL has initialization problems in the
redundancy sense that effects the total codelength, which we solve
it  by proposing another predictive approach that is called the sufficient
statistic method, in which using sufficient statistics $\mathbf{t}$,
the distribution of the parameters $\boldsymbol{\hat{\theta}}(\mathbf{t})$
are estimated and applied to calculate the probability of the next
sample $P_{\boldsymbol{\hat{\theta}}(\mathbf{t})}^{ss}(x_{i+1}|\mathbf{x}_{i})$.
To compare our sufficient statistic method and the ordinary predictive
approach, assume the model is $\mathcal{N}(\mu,\sigma^{2})$, then\vspace{-0.2in}

\begin{align}
P^{op}(x_{n+1}|\mathbf{x}_{n}) & =\frac{1}{\sqrt{2\pi S_{n}^{2}}}\exp\left(-\frac{1}{2S_{n}^{2}}\left(x_{n+1}-\widehat{\mu_{n}}\right)^{2}\right)\label{eq:OP}\\
P^{ss}(x_{n+1}|\mathbf{x}_{n}) & =\sqrt{\frac{n}{\pi\left(n+1\right)}}\frac{\Gamma\left(\frac{n+1}{2}\right)}{\Gamma\left(\frac{n}{2}\right)}\frac{\left[nS_{n}^{2}\right]^{\frac{n}{2}}}{\left[\left(n+1\right)S_{n+1}^{2}\right]^{\frac{n+1}{2}}}\label{eq:SS}
\end{align}
where $\widehat{\mu_{n}}=\frac{1}{n}\sum_{i=1}^{n}x_{i}$ and $S_{n}^{2}=\frac{1}{n-1}\sum_{i=1}^{n}\left(x_{i}-\widehat{\mu_{n}}\right)^{2}$.
Note that for convenience, we drop the subscript $\boldsymbol{\hat{\theta}}(\mathbf{x}_{i})$
in probabilities. Even though both ordinary predictive MDL and sufficient
statistic methods have distinct behavior in their initialization performance,
they both achieve the same asymptotic code length. In section \ref{sec:Coding}
we'll see how this predictor are going to be used in each node of
pattern-tree to calculate coding distributions.

\section{Pattern tree: a model for heart rate variability}

As it was mentioned earlier in the introduction, in \cite{Host13ITW}
we  discovered some patterns in HRV that are signs of particular heart
arrhythmias. Due to quantization of the data, our discovery was limited,
therefore we are motivated to extend the result to the real-valued
time series of HRV and this requires a model for HRV signal. Even
though there are many nonlinear autoregressive processes and switching
state-space models for HRV, Costa \emph{et al }\cite{GaussMixModel}
showed that all of those methods give a rise to a multimodal distribution
for HRV and they concluded that the best is using Gaussian mixture
model. Existence of premature beats (heart beats that happen very
early, before heart contraction happens) in the HRV data that we have
used in \cite{Host13ITW} was indicative of the same fact at the rudimentary
level. 

Consider a time series of heart rate measurements $\mathbf{x}_{n}=\left\{ x_{1},x_{2},\cdots,x_{n}\right\} $
and define $z_{n}$ as the state variable at time $n$ with $K$ possibilities
($K$ is the number of Gaussian distributions in the mixture model).
Let $\pi_{k}=\Pr\left\{ z_{n}=k\right\} $ for all $1\leq k\leq K$,
then based on Costa's model\emph{ }\cite{GaussMixModel} we have \vspace{-0.05in}
\begin{align}
P\left(x_{n+1}|\mathbf{x}_{n}\right) & =\sum_{z_{n}}P(z_{n})P(x_{n+1}|\mathbf{x}_{n},z_{n})\nonumber \\
 & =\sum_{k=1}^{K}\pi_{k}P_{\boldsymbol{\theta}_{k}}(x_{n+1}|\mathbf{x}_{n})\label{eq:HrvModel}
\end{align}
where $\mathbf{\pi}=\left\{ \pi_{k}\right\} _{k=1}^{K}$ satisfies
$0\leq\pi_{k}\leq1$ and $\sum_{k=1}^{K}\pi_{k}=1$, and $P_{\boldsymbol{\theta}_{k}}(x_{n+1}|\mathbf{x}_{n})$
is a Gaussian distribution with $\boldsymbol{\theta}_{k}=\left\{ \mu_{k},\sigma_{k}^{2}\right\} $.

The model (\ref{eq:HrvModel}) and the biological reason behind mixture-modeling
of HRV made us consider a more complex and more exhaustive model for
this process, a tree model. This can be explained via a simple example.
Assume we have a binary tree with depth one, this tree partitions
the samples in the root node by comparing it with the previous sample
and based on having a rise or a drop, it will be assigned to one of
the child nodes. Now suppose this tree is used to divide the samples
in a HRV signal that contains premature beats. Note that the histogram
of this HRV signal will be a mixture of two Gaussian distributions
with separated means and different variances. Since the time difference
between the two consecutive normal heart beats is (relatively) much
larger than the time interval between a normal heart beat and a premature
one, the depth-one tree will separate all the premature beats from
normal ones (however some normal beats can be classified as premature,
since depth-one tree is too simple). By letting this tree structure
to be deeper, more complex arrhythmias such as tachycardia, flutter
and fibrillation will be separated from normal beats; this could also
lead to new discoveries of heart abnormalities, which is the real
goal of our work. Since the same tree structure is used in CTW and
its application led to data discovery (\cite{HostSabetiWalton16IEEEtrans,Host13ITW}),
here we want to extend the same concept to develop the \emph{pattern
tree weighting (PTW)} for real-valued data. This proposed method uses
the binary data pattern as context and apply  predictive MDL for real-valued
samples to compute the coding distribution.

\subsection{\label{sec:PTW}Pattern tree as an extension of the context tree}

The context tree has been shown to be a powerful method to compute
an appropriate coding distribution \cite{WillemsAl95,WillemsAl97}.
A context of the binary source symbol $y_{n}$ is a suffix of the
semi-infinite sequence $\cdots,y_{n-2},y_{n-1}$ that precedes it.
The context tree consists of nodes corresponding to each context up
to certain depth. A pattern tree of depth $D$ has the same structure
of a context tree of the same depth, but since we are interested to
design it for real-valued data, the way it splits the source sequence
is different. Suppose at time $n$, the last $D$ samples of the real-valued
source sequence ($x_{i}\in\mathbb{R}$) are $x_{n-D},\cdots,x_{n-2},x_{n-1}$.
After putting $x_{n}$ in the root node $\eta_{1}=\lambda$, the way
we assign $x_{n}$ to any of the root's children ($\eta_{2}$ and
$\eta_{3}$) at depth one is based on comparing $x_{n}$ and $x_{n-1}$,
for instance if $x_{n-1}>x_{n}$ we assign $x_{n}$ to node $\eta_{2}$.
Next the way we assign $x_{n}$ to $\eta_{2}$'s children ($\eta_{4}$
and $\eta_{5}$) at depth two is based on comparing $x_{n-1}$ and
$x_{n-2}$ and we keep on doing this until we reach the maximum depth
$D$. As it can be seen, at time $n$, connecting the nodes that are
assigned by $x_{n}$ illustrates a pattern that shows the fluctuation
of the $(D+1)$-most recent source samples from $x_{n-D}$ to $x_{n}$,
and that's the reason we call it a pattern tree. Here is an example
to show how the source sequence is portioned by the pattern tree.\vspace{-0.1in}

\begin{example}
\label{exa:Split}Consider a tree with depth $D=3$ and suppose that
the source generated the sequence {[}1.7 0.6 -2.6 -0.8 0.7 7.1 5.5
-2.7 6 1.4{]} while the past sequence was {[}... -0.9 0.1 -0.4{]}.
Then the sequence is partitioned by the pattern tree, see Fig. \ref{fig:SplitProbability}
(for now disregard the probabilities in the Fig. \ref{fig:SplitProbability}).\vspace{-0.15in}
\end{example}

\subsection{\label{sec:Coding}Coding for an unknown tree source}

In this section we describe how to employ the predictive MDL of section
\ref{sec:Predictive-MDL} to estimate the probability in each node.
Here we want to use the same concept of the weighted coding distribution
in the context tree, adapted to the real-value case. Therefore to
each node $\eta$ we will assign a sequential predictive distribution
that only depends on the data samples observed by this particular
node. In fact, the weighted coding distribution of the pattern tree
will be exactly like the context tree, but with this difference that
instead of KT-estimator which is designed for binary data, we use
predictive MDL estimators (e.g., for the case of Gaussian see (\ref{eq:OP})
or (\ref{eq:SS}) ). 

\begin{figure}[tbh]
\includegraphics[width=3.5in,height=2.3in]{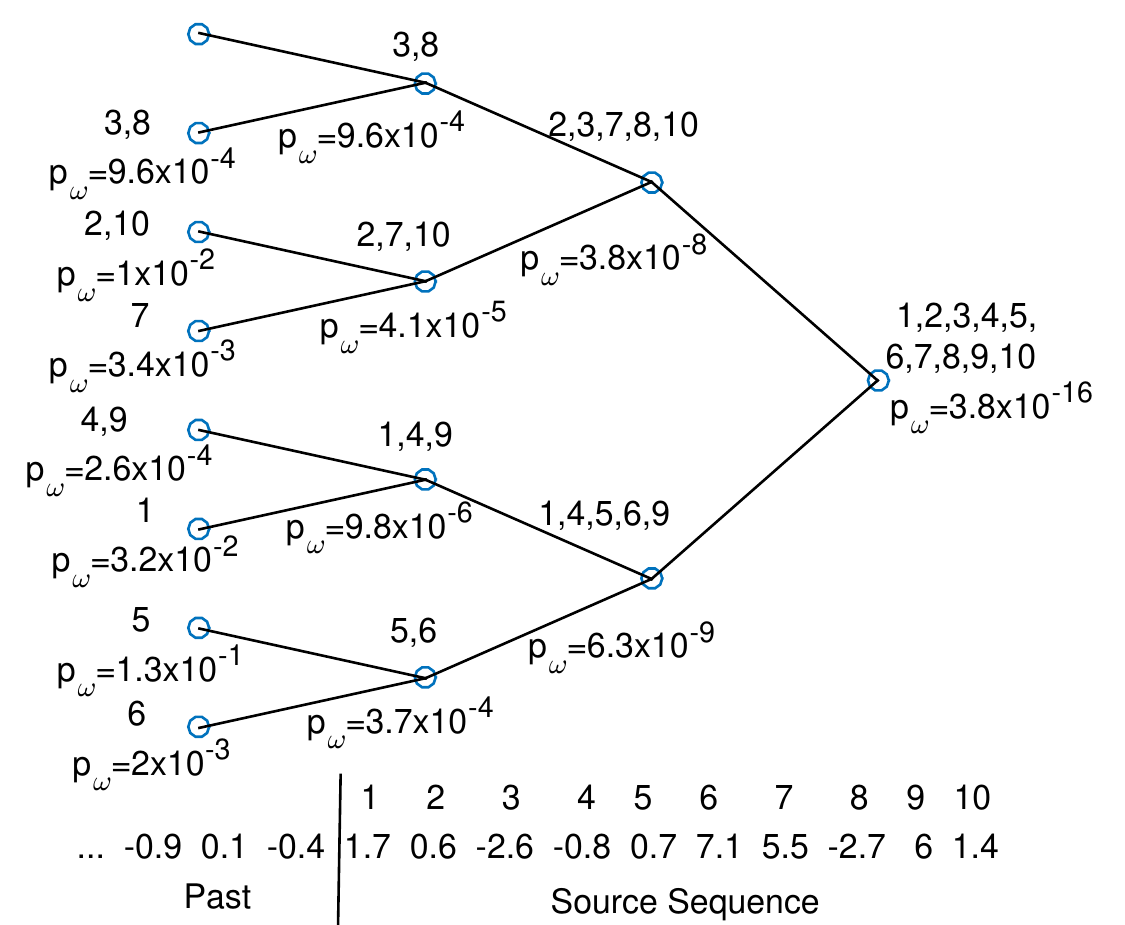}\vspace{-0.15in}

\caption{\label{fig:SplitProbability}The source sequence of Example \ref{exa:Split}
is split up by PTW. In Example \ref{exa:SplitProbability} the weighted
probabilities are then calculated.}

\vspace{-0.1in}
\end{figure}

From now on, we assume for a tree with depth $D$, the initial context
of past sequence $x_{1-D},\cdots,x_{0}$ is available. At any time
$n$ for each node $\eta_{i}$ suppose the vectors $t_{n}^{\eta_{i}}=\left\{ t<n\::\:x_{t}\in\eta_{i}\right\} $
and $\mathbf{x}_{n}^{\eta_{i}}=\left\{ x_{t}\::\:t\in t_{n}^{\eta_{i}}\right\} $
are the set of all the time indexes of the already observed data samples
by that node and its corresponding data samples, respectively. Clearly
for every internal node we have $t_{n}^{\eta_{i}}=t_{n}^{\eta_{2i}}\cup t_{n}^{\eta_{2i+1}}$,
i.e., the set of all the time indexes of the already observed data
samples by the parent node $\eta_{i}$ is the union of the set of
all the time indexes of the already observed data samples by its children
(see Fig. \ref{fig:SplitProbability}), and similarly $\mathbf{x}_{n}^{\eta_{i}}=\mathbf{x}_{n}^{\eta_{2i}}\cup\mathbf{x}_{n}^{\eta_{2i+1}}$.
Ergo the weighted probability in each internal node is

\begin{align*}
P_{\omega,\eta_{i}}(x_{n+1}|\mathbf{x}_{n}^{\eta_{i}}) & =\frac{1}{2}P_{\eta_{i}}(x_{n+1}|\mathbf{x}_{n}^{\eta_{i}})\\
 & +\frac{1}{2}P_{\eta_{2i}}(x_{n+1}|\mathbf{x}_{n}^{\eta_{2i}})P_{\eta_{2i+1}}(x_{n+1}|\mathbf{x}_{n}^{\eta_{2i+1}})
\end{align*}
where $P_{\eta_{i}}(x_{n+1}|\mathbf{x}_{n}^{\eta_{i}})$ is the predictive
distribution (e.g., (\ref{eq:OP}) and (\ref{eq:SS})) at node $\eta_{i}$.
So in each node, the parameter of the model is estimated based on
the observed samples at that particular node. One issue here can be
in initialization, for instance for a Gaussian source, at least each
node needs two samples to estimate both mean and variance. This can
be resolved by estimating the initial  predictive distribution parameters
based on the first $D$ samples of data in the context. Finally the
weighted probability in each internal node will be (see (\ref{eq:Pw})
at the top of the next page).

\begin{algorithm*}[tbh]
\vspace{-0.1in}
\begin{align}
P_{\omega,\eta_{i}}(x_{n+1}|\mathbf{x}_{n}^{\eta_{i}}) & =\begin{cases}
\frac{1}{2}P_{\eta_{i}}(x_{n+1}|\mathbf{x}_{n}^{\eta_{i}})+\frac{1}{2}P_{\eta_{2i}}(x_{n+1}|\mathbf{x}_{n}^{\eta_{2i}})P_{\eta_{2i+1}}(x_{n+1}|\mathbf{x}_{n}^{\eta_{2i+1}}) & \qquad\eta_{i}\,:\:internal\:node\\
\\
P_{\eta_{i}}(x_{n+1}|\mathbf{x}_{n}^{\eta_{i}}) & \qquad\eta_{i}\,:\:leaf\:node
\end{cases}\label{eq:Pw}
\end{align}
\vspace{-0.1in}

\end{algorithm*}
As an example for predictive distribution, if the source sequence
is generated according to $\mathcal{N}(\mu,\sigma^{2})$ where both
$\mu$ and $\sigma^{2}$ are unknown, then using ordinary predictive
MDL of (\ref{eq:OP}) we have\vspace{-0.2in}

\begin{align}
P_{\eta_{i}}(x_{n+1}|\mathbf{x}_{n}^{\eta_{i}}) & =\frac{1}{\sqrt{2\pi S_{n}^{2}}}\exp\left(-\frac{1}{2S_{n}^{2}}\left(x_{n+1}-\widehat{\mu_{n}}\right)^{2}\right)\label{eq:OPestimator}
\end{align}
where $\widehat{\mu_{n}}=\frac{1}{|\mathbf{x}_{n}^{\eta_{i}}|}\sum_{x_{i}\in\mathbf{x}_{n}^{\eta_{i}}}x_{i}$
and $S_{n}^{2}=\frac{1}{|\mathbf{x}_{n}^{\eta_{i}}|-1}\sum_{x_{i}\in\mathbf{x}_{n}^{\eta_{i}}}\left(x_{i}-\widehat{\mu_{n}}\right)^{2}$.
The following example calculates the weighted probability of the source
sequence of Example \ref{exa:Split} using Gaussian MDL predictor
(\ref{eq:SS}):\vspace{-0.1in}

\begin{example}
\label{exa:SplitProbability}Back to the Example \ref{exa:Split},
the weighted probabilities $P_{\omega}$ using our sufficient-statistic-based
predictor (\ref{eq:SS}) as the estimator for Gaussian distribution
are depicted in the pattern tree in Fig. \ref{fig:SplitProbability}.\vspace{-0.2in}
\end{example}

\vspace{-0.2in}

\subsection{Updating the pattern tree and complexity}

Suppose a pattern tree of depth $D$ already seen the source sequence
$x_{1},\cdots,x_{n-1},x_{n}$, now we want to see how complex the
pattern tree evolves when the next sample $x_{n+1}$ is going to be
processed. This is done by updating all the nodes in only one path
of length $D$ from the root to a leaf node and this path is determined
by the procedure explained in section \ref{sec:PTW}. Then the updating
steps in all the $D+1$ nodes in the evolving path are: (i) updating
the parameter estimation, (ii) updating predictive probabilities $P_{\eta_{i}}(x_{n+1}|\mathbf{x}_{n}^{\eta_{i}})$
and (iii) updating the weighted probabilities $P_{\omega,\eta_{i}}(x_{n+1}|\mathbf{x}_{n}^{\eta_{i}})$.
The following example which is the continuation of the series of examples
shows how the source sequence in Example \ref{exa:SplitProbability}
is updated for new source sample.\vspace{-0.1in}

\begin{example}
Suppose that the source has already generated the sequence of Example
\ref{exa:Split}. This resulted the weighted pattern tree in Fig.
\ref{fig:SplitProbability}.\vspace{-0.2in}
\end{example}
\begin{figure}[tbh]
\includegraphics[width=3.5in,height=2.2in]{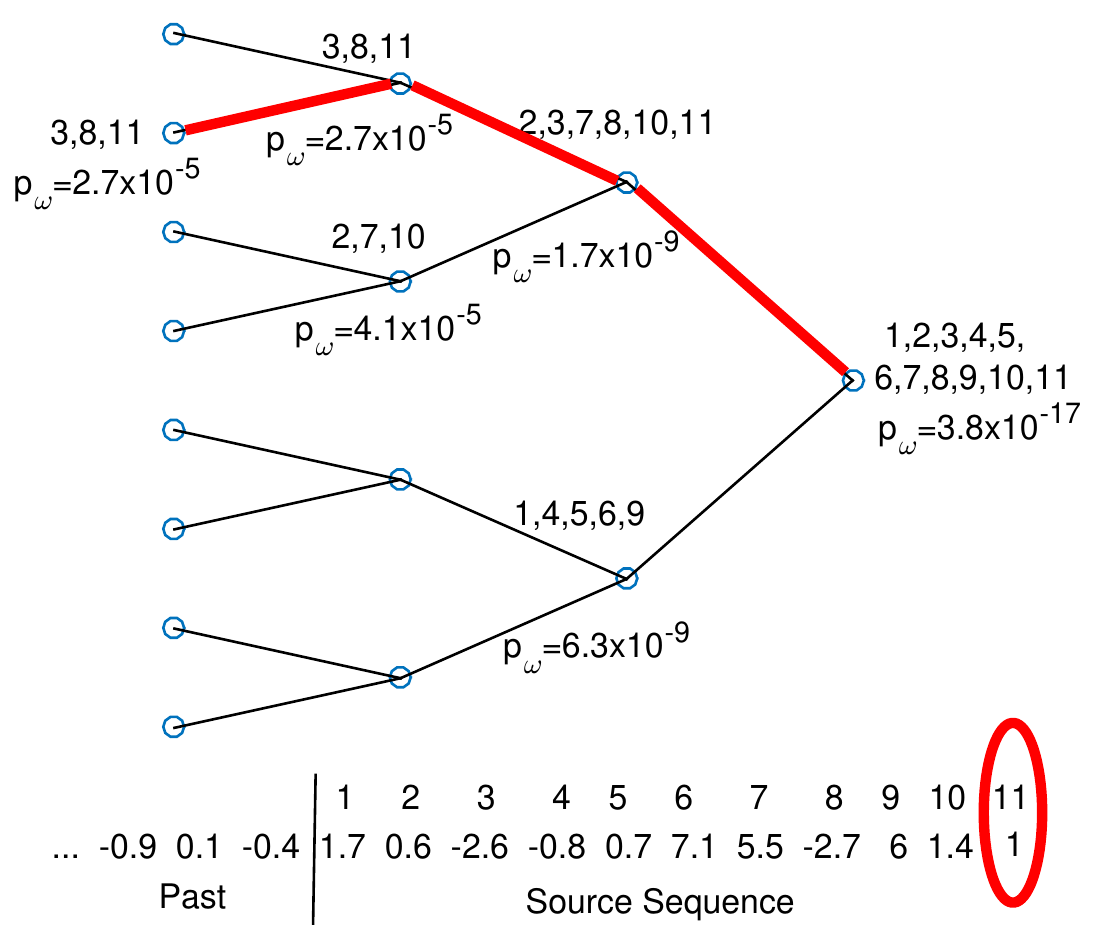}\vspace{-0.15in}

\caption{\label{fig:Update}Updated path of the weighted pattern tree of the
source sequence in Example \ref{exa:Split} followed by a 1.}

\vspace{-0.1in}
\end{figure}
Note that by assuming a parametric model (here, a Gaussian model),
there would be no more need to store all the data samples in each
node; instead, only parameters should be updated and stored in the
nodes. 

\vspace{-0.2in}

\section{Atypicality using PTW}

Suppose $\mathcal{P}=\left\{ P_{\boldsymbol{\theta}}\left(\mathbf{x}_{n}\right):\:\boldsymbol{\theta}\in\Theta\subset\mathcal{\mathbb{R}}^{k}\right\} $
is a family of parametric model class that can be assigned to the
data. In terms of coding, Definition \ref{atypdef.thm} can be stated
in the following form\vspace{-0.1in}
\[
C(\mathbf{x}|\mathcal{P})-C(\mathbf{x})>0
\]
Here $C(\mathbf{x}|\mathcal{P})$ is the code length of $\mathbf{x}$
encoded with the optimum coder according to the typical law (known
parameters $\boldsymbol{\theta}$), and $C(\mathbf{x})$ is $x$ encoded
'in itself.' As argued in \cite{HostSabetiWalton16IEEEtrans}, we
need to put a 'header' in atypical sequences to inform the encoder
that an atypical encoder is used. We can therefore write $C(\mathbf{x})=\tau+\tilde{C}(\mathbf{x})$,
where $\tau$ is the number of bits for the 'header,' and $\tilde{C}(\mathbf{x})$
is the number of bits used for encoding the data itself. For encoding
the data in itself an obvious solution is to use a universal source
coder. We have therefore chosen to use the PTW algorithm.\vspace{-0.1in}

\subsection{\label{Typical.sec}Typical Encoding and Training}

\begin{figure}[tbh]
\includegraphics[width=3.5in,height=2.2in]{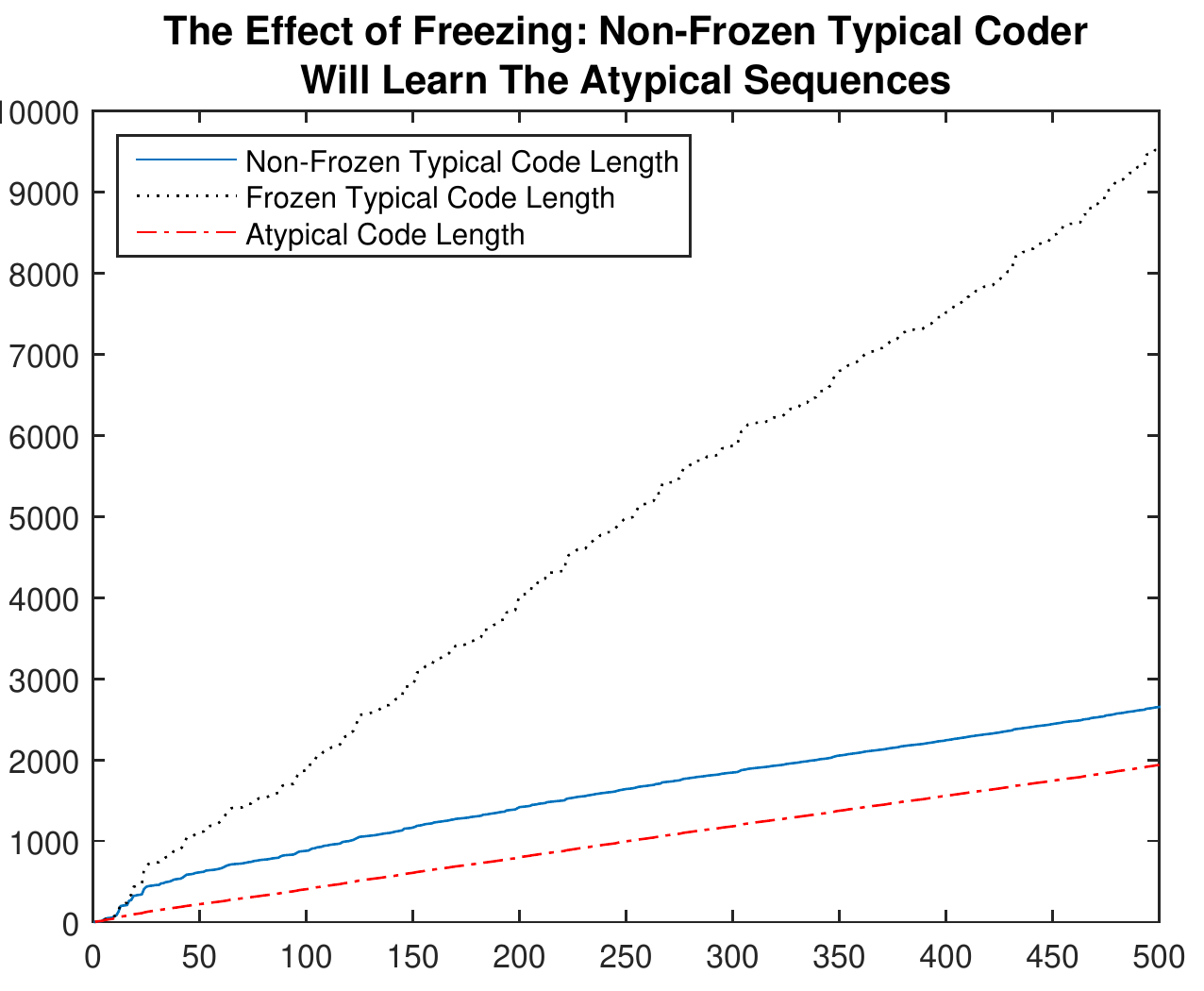}\vspace{-0.15in}

\caption{\label{fig:FreezingEffect}The importance of freezing source coding
when testing for atypicality. }

\vspace{-0.1in}
\end{figure}

In Definition \ref{atypdef.thm} we have assumed that parameters $\boldsymbol{\theta}$
of the typical model of data is exactly known, however, in many cases
the parameters are not known exactly. Let us assume we are given a
single long sequence $\mathbf{t}$ for training \textendash{} rather
than the parameters $\boldsymbol{\theta}$ in model $\mathcal{P}$
\textendash{} and based on this we need to encode a sequence $\mathbf{x}$.
To understand what this means, we have to realize that when $\mathbf{x}$
is encoded according to $C(\mathbf{x}|\mathcal{P})$ with a known
$\mathcal{\boldsymbol{\theta}}$, the coding probabilities are \emph{fixed};
they are not affected by $\mathbf{x}$. This is an important part
of Definition \ref{atypdef.thm} that reacts to 'outliers,' data that
does not fit the typical model. But the issue with universal source
coders is that they often easily adapts to new types of data, a desirable
property of a good universal source coder, but problematic in light
of the above discussion. We therefore need to 'freeze' the source
coder, for example by not updating the dictionary. However, because
the training data is likely incomplete as discussed above, the freezing
should not be too hard. This issue is precisely described in \cite{HostSabetiWalton16IEEEtrans}
and due to page limitation we don't go over it here, but with a simulation
we show why freezing the encoder is essential in implementing atypicality:
 The PTW algorithm trained with a Gaussian process with $\sigma^{2}=4$
and then tested with another Gaussian source with same mean and different
variance $\sigma^{2}=1$. It can be seen in Fig. \ref{fig:FreezingEffect}
that the non-frozen algorithm learn the statistic and behavior of
the new source; however, the codelength using frozen algorithm keeps
increasing and that's a property we are interested in. \vspace{-0.15in}

\vspace{-0.2in}

\subsection{\label{subsec:AtypSubseq}Atypical subsequences}

 Let $\mathcal{X}(n,l)=(x_{n},x_{n+1},\ldots,x_{n+l-1})$ be a subsequence
of $\{\mathbf{x}_{n},n=0,\ldots,\infty\}$ that we want to test for
atypicality. As mentioned earlier, the start of a sequence needs to
be encoded as well as the length. Additionally the code length is
minimized over the maximum depth $D$ of the context tree. The atypical
code length is then given by 
\begin{align*}
L_{A}(\mathcal{X}(n,l))= & \min_{D}\left(-\log P_{\omega,\lambda}(D)+\log^{*}D\right)+\log^{*}l
\end{align*}
\emph{except} for the $\tau$. Here $P_{\omega,\lambda}(D)$ denotes
the probability at the root of the pattern tree of depth $D$. For
typical coding we use  the algorithm in Section \ref{Typical.sec}
when the parameters are not known;  let $L(n)$ be the codelength
for the sequence $x_{0},\ldots,x_{n}$. Then we can put $L_{T}(\mathcal{X}(n,l))=L(n+l-1)-L(n)$.
We need to test every subsequence of every length, that is, we need
to test subsequences $\mathcal{X}(n,l)$ for every value of $n$ and
$l$. For atypical coding this means that a new PTW algorithm needs
to be started at every sample time. So, if the maximum sequence length
is $L$, $L$ separate PTW trees need to maintained at any time. These
are completely independent, so they can be run on parallel processors.
The result is that for every bit of the data we calculate 
\begin{align}
\Delta L(n) & =\min_{l}\left\{ L_{A}(\mathcal{X}(n,l))-L_{T}(\mathcal{X}(n,l))\right\} \label{DeltaLn.eq}
\end{align}
and we can state the atypicality criterion as $\Delta L(n)<-\tau$. 

\vspace{-0.2in}

\section{Anomaly Detection}

In \cite{HostSabetiWalton16IEEEtrans} we have used CTW for anomaly
detection in binary data and we got unique results. In order to verify
the performance of our algorithm for real-valued data, here we want
to use PTW as encoder in atypicality framework for same purpose. \vspace{-0.2in}

\subsection{Comparison with an alternative method}

\begin{figure}[tbh]
\includegraphics[width=3.5in,height=1.9in]{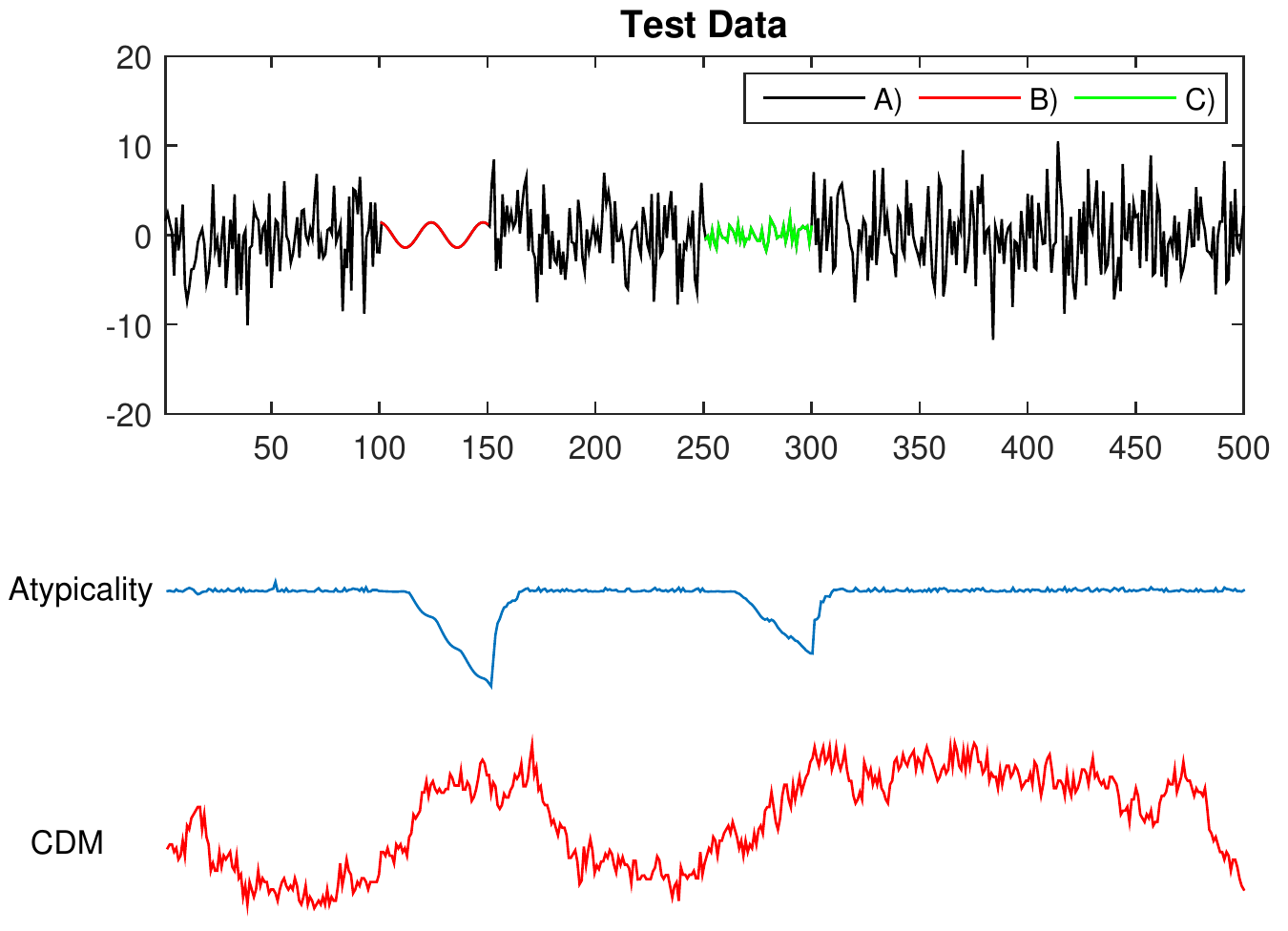}\vspace{-0.15in}

\caption{\label{fig:Comparison}A comparison of two different methods. For
our proposed method, the $\triangle L$ of equation (\ref{DeltaLn.eq})
has been plotted. A) Gaussian process with zero mean and $\sigma^{2}=4$,
exactly the same as training (typical) data, B) A sinusoidal segment
C) A segment of Gaussian process with zero mean and $\sigma^{2}=1$
.}

\vspace{-0.1in}
\end{figure}

In \cite{KeoghAl04} authors came up with the Compression-based Dissimilarity
Measure (CDM) that uses the dissimilarity measure for anomaly detection.
They showed by that their proposed algorithm outperformed other methods.
Here we want to compare our algorithm with CDM on a simulated data.
In this experiment the training sequence is a Gaussian process with
zero mean and $\sigma^{2}=4$, and the test sequence have the same
statistics as the training data but with two episodes of anomalies
embedded in it: a sinusoidal segment (red part in Fig. \ref{fig:Comparison})
and a segment of Gaussian process with zero mean and $\sigma^{2}=1$
(green part in Fig. \ref{fig:Comparison}). As can be verified from
the figure, our PTW-based atypicality found both of the anomalies
in the data; however, CDM only detected the sinusoidal pattern. \vspace{-0.2in}

\vspace{-0.2in}

\subsection{Detection of Paroxysmal Atrial Fibrillation (PAF)}

\begin{figure}[tbh]
\includegraphics[width=3.5in,height=2.2in]{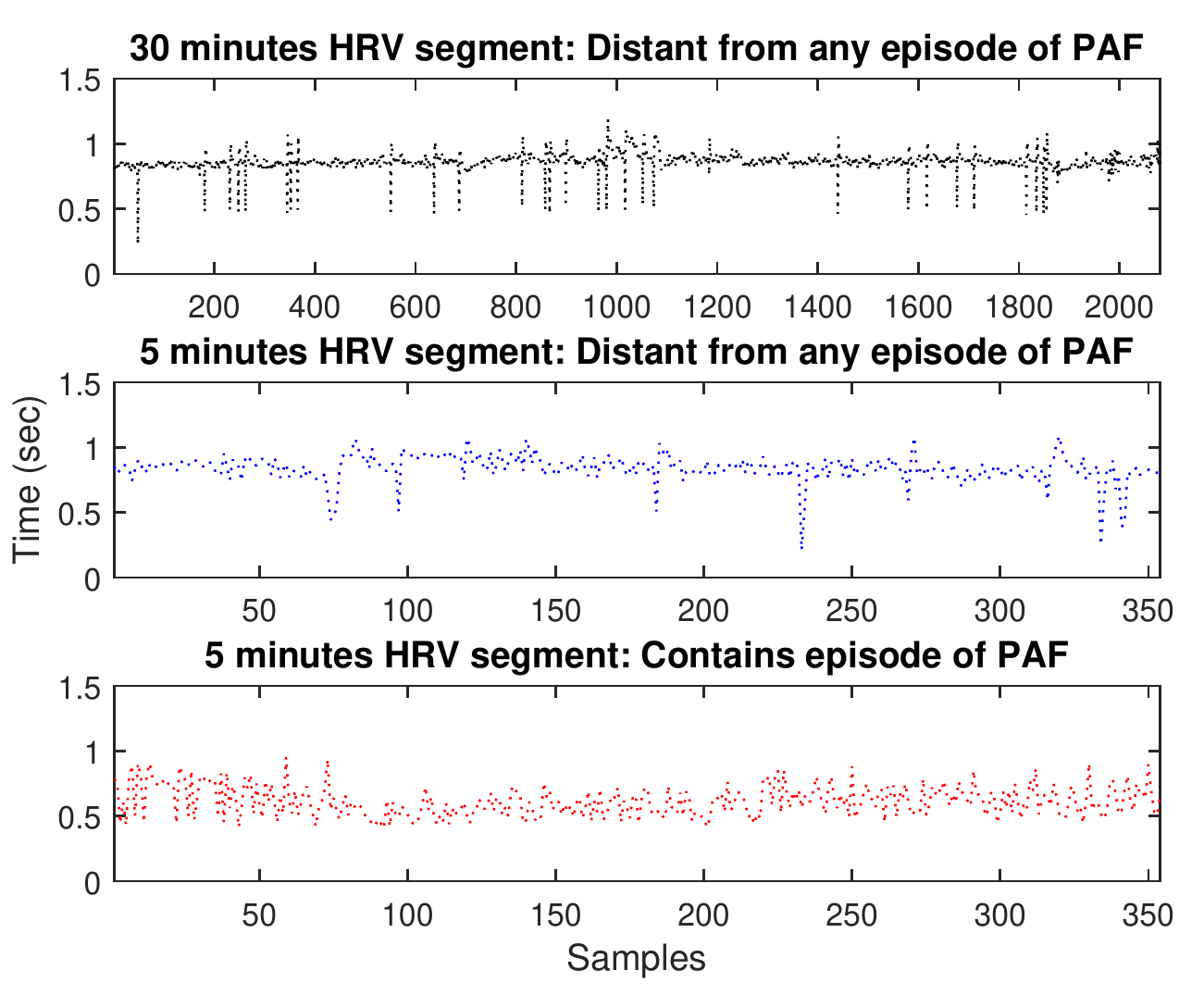}\vspace{-0.15in}

\caption{\label{fig:PAF}HRV signal of a PAF subject in the database}

\vspace{-0.1in}
\end{figure}

Paroxysmal Atrial Fibrillation (PAF) is an irregular, often rapid
heart rate that commonly causes poor blood flow, which may lead to
severe consequences. PhysioNet \cite{Goldbergere215} has provided
a database for this arrhythmia \cite{moody2001predicting}. Part of
this database includes 30-minute ECG records of different subjects
(who have PAF) during a period that is distant from any episode of
PAF. Each 30-minute record is then followed by two 5-minute record
of the same subject, one of which contains an episode of PAF, and
the other one has no such an episode.  Fig. \ref{fig:PAF} shows
HRV signal for a subject in the database.\vspace{-0.01in}

 In our experiment, for each subject we train the PTW with HRV of
30-minute records and freeze it, then it was used as encoder for its
two 5-minute segments. Our goal was to detect the 5-minute record
with PAF episode, since we believed using the PTW trained on 30-minute
records, smaller codelength is needed to encode the 5-minute record
without PAF episode. After applying the same procedure on all the
25 subjects, we were able to detect all the 5-minute record containing
PAF episode correctly, with 100 accuracy (consequently due to data
structures, zero false alarm), in fact we have improved the best results
of other researches on the same database about 10 percent \cite{sabeti2012prediction,thong2004prediction}.
Again, as we discussed in section \ref{Typical.sec} it verifies the
importance of freezing in our notion of atypicality.\vspace{-0.2in}

\subsection{Detection of anomaly in Holter Monitoring}

As we mentioned earlier, the ultimate goal of atypicality is to find
``unknown unknowns'' in Big Data. One example can be the data that
achieved by Holter Monitoring, i.e., a continuous tape recording of
a patient's ECG for 24 hours. For such a data we used the MIT-BIH
Normal Sinus Rhythm Database (nsrdb) which is provided by PhysioNet
\cite{Goldbergere215}.  Even though the subjects included in this
database were found to have had no significant arrhythmias, there
exist many arrhythmic beats and patterns to look for. We want to apply
our algorithm to find interesting parts of the data in the dataset.

\begin{figure}[tbh]
\includegraphics[width=3.4in,height=2.2in]{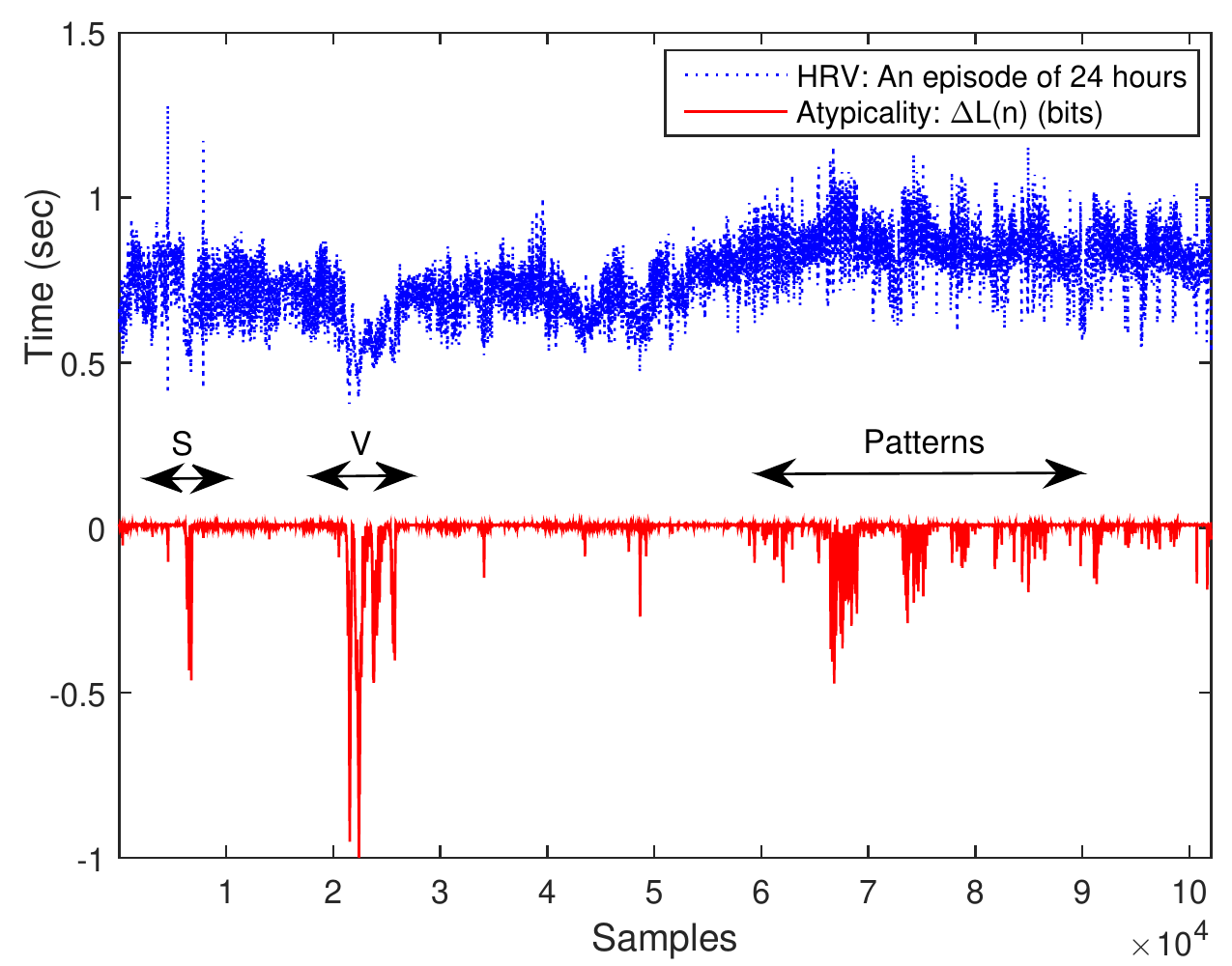}\vspace{-0.15in}

\caption{\label{fig:HRV}Atypicality using PTW for Holter Monitoring HRV: ``S''
stands for supraventricular arrhythmia and ``V'' stands for ventricular
contraction }

\vspace{-0.1in}
\end{figure}

 Since the data is assumed to be ``Normal Sinus Rhythm,'' we trained
and froze our typical PTW with Gaussian process of the same mean and
variance of the data. Using the process of the section \ref{subsec:AtypSubseq},
we managed to find interesting results on the dataset. As an example,
we provide Fig. \ref{fig:HRV}. An can be seen in the figure, for
that particular data we came up with three major atypical subsequences:
the first one is denotes by ``S,'' second one is marked with ``V''
and the last one as ``pattern.'' After looking up in the annotation
file which is provided for the HRV, the ``S'' group corresponds
to onset of some supraventricular beats and the ``V'' groups was
the result of ventricular contraction; however, there were no label
for the segment that we call ``pattern'' in the HRV annotations.
Looking closer in the data shows existence of some repetitive patterns
that was not even seen by the cardiologist who annotated the HRV data.
Taking a deeper look into ECG (not HRV) annotation shows that those
patterns are happening at the same time that there are either some
isolated QRS-like artifact or signal quality change \cite{Goldbergere215}.
This shows our algorithm was able to find something that was missed
by the eye of the expert who was only looking into the HRV data without
considering the ECG signals; nevertheless, it could be the sign of
some heart malfunction after the recording was over. \vspace{-0.15in}

\vspace{-0.1in}

\bibliographystyle{IEEEtran}
\bibliography{Coop06,Sensor,ahmref2,Coop03,BigData,Underwater,CDMA,combined,ECGandHRV}

\end{document}